\documentclass[letterpaper, 10 pt, conference]{ieeeconf}  

\IEEEoverridecommandlockouts                              

\usepackage{graphicx} 
\usepackage{epsfig} 
\usepackage{mathptmx} 
\usepackage{times} 
\usepackage{amsmath} 
\usepackage{amssymb}  
\usepackage{booktabs} 
\usepackage{multirow} 
\usepackage{algorithm}
\usepackage{algorithmic}
\usepackage{algorithm}
\usepackage{algorithmic}
\usepackage{xcolor,color}
\usepackage{amsmath}
\usepackage{tabularx}
\usepackage{multirow}

\title{MACE: Mixture-of-Experts Accelerated Coordinate Encoding for Large-Scale Scene Localization and Rendering
}

\author{Mingkai Liu$^{1}$, Dikai Fan$^{1}$, Haohua Que$^{2}$, Haojia Gao$^{3}$, Xiao Liu$^{1}$, Shuxue Peng$^{1}$, Meixia Lin$^{1}$, \\Shengyu Gu$^{1}$, Ruicong Ye$^{4}$, Wanli Qiu$^{4}$, Handong Yao$^{2}$, Ruopeng Zhang$^{5}$, Xianliang Huang$^{1,*}$  
\thanks{$^{1}$PICO, ByteDance Inc. }%
\thanks{$^{2}$University of Georgia, Athens, USA.}%
\thanks{$^{3}$Beijing University of Technology, Beijing, China. }%
\thanks{$^{4}$Peking University, Beijing, China.}%
\thanks{$^{5}$Chongqing Vocational Institute of Engineering, Chongqing, China.}%
\thanks{$^{*}$Corresponding author: Xianliang Huang (email: huangxianliang@bytedance.com).}
}

\begin{document}

\maketitle
\thispagestyle{empty}
\pagestyle{empty}

\begin{abstract}
	Efficient localization and high-quality rendering in large-scale scenes remain a significant challenge due to the computational cost involved. While Scene Coordinate Regression (SCR) methods perform well in small-scale localization, they are limited by the capacity of a single network when extended to large-scale scenes.
	To address these challenges, we propose the Mixed Expert-based Accelerated Coordinate Encoding method (MACE), which enables efficient localization and high-quality rendering in large-scale scenes. Inspired by the remarkable capabilities of MOE in large model domains, we introduce a gating network to implicitly classify and select sub-networks, ensuring that only a single sub-network is activated during each inference.
	Furtheremore, we present Auxiliary-Loss-Free Load Balancing (ALF-LB) strategy to enhance the localization accuracy on large-scale scene.
	Our framework provides a significant reduction in costs while maintaining higher precision, offering an efficient solution for large-scale scene applications.
	Additional experiments on the Cambridge test set demonstrate that our method achieves high-quality rendering results with merely 10 minutes of training.

\end{abstract}

\section{INTRODUCTION} \label{sec::intro}
Large-scale scene localization and rendering holds significant value in computer vision, which involves recovering the camera pose from a sequence of images and reconstructing visually and structurally complete scenes.
However, this task faces considerable challenges stemming from the fundamental trade-off between efficiency, computation burden, and accuracy. Providing robust solutions for large-scale localization and rendering has a direct impact on key applications such as Augmented Reality~\cite{sarlin2022lamar,baker2024localization}, autonomous driving~\cite{jiang2023vad, zhu2024scene}, and robotic navigation~\cite{yang2019sanet, liu2025embodied}.

\begin{figure}[t]
	\centering
	\scalebox{1}{\includegraphics[width=1.0\linewidth]{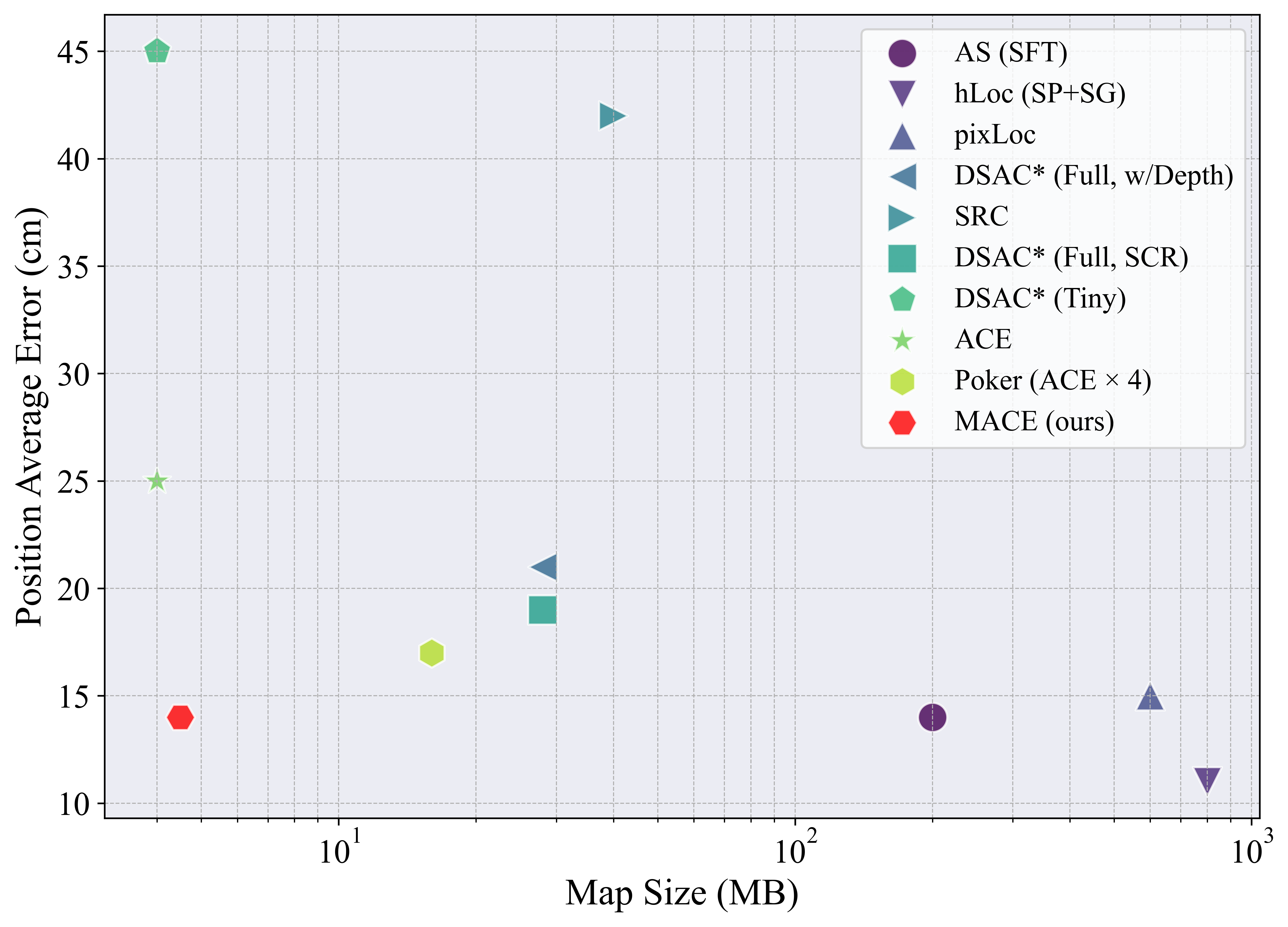}}

	\caption{
		\textbf{Map Size vs Position Average Error.} Comparison with several counterparts on the Cambridge dataset demonstrates that our method reduces activation weight by 72\% vs. Poker (ACE $\times$ 4) while maintaining low localization errors.
	}
	\label{fig:problem}
\end{figure}


Mainstream localization methods can be divided into two categories: structure-based and Scene Coordinate Regression (SCR) approaches.
Structure-based methods utilize SfM to reconstruct 3D point clouds and estimate camera poses via 2D–3D descriptor matching and PnP solvers. However, they incur high computational and storage costs, especially in large-scale scenes due to the need to store dense visual descriptors.
In contrast, SCR methods encode scene maps implicitly via deep neural networks, directly regressing 3D coordinates from 2D features without explicit descriptor matching. They perform well in small scenes~\cite{shotton2013scene,brachmann2023accelerated} with fast training times. However, their scalability is limited, as a single network struggles to capture global information in large-scale environments.

Existing methods~\cite{budvytis2019large,brachmann2019expert} address large-scale scene localization by employing scene clustering and multi-subnetwork training, where each sub-network specializes in a specific sub-region. While these methods improve accuracy through parallel inference and optimal sub-network selection, it incurs high computational costs and suffers from performance degradation due to suboptimal clustering, limiting its practical applicability. Other approaches~\cite{li2020hierarchical,jiang2025r} leverages global graph encoding and data augmentation, along with depth-aware losses to enhance scalability, but often introduce additional network complexity and preprocessing overhead.

On the other hand, a visually realistic and complete reconstruction of the environment is essential for many applications, going beyond the sparse point sets typically used for localization.
In most VR/AR applications, users often need to quickly capture specific real-world views and reproduce them with high quality.
Existing methods like SCR and SfM-based approaches typically produce sparse, textureless point clouds that lack details, especially in low-texture regions. They cannot be directly used in application development platforms like Unity for developers to build AR or robotics applications.
Recently, 3DGS has become a mainstream solution for 3D scene representation due to its excellent real-time rendering capability and high-fidelity visual effects.
However, traditional 3DGS heavily relies on accurate point cloud priors provided by methods such as SfM, limiting its practicality in large-scale scene rendering.


To address the aforementioned challenges of large-scale scene localization and rendering, we propose an innovative framework \textbf{M}ixture-of-experts \textbf{A}ccelerated \textbf{C}oordinate \textbf{E}ncoding termed \textbf{MACE}.
Different with the Mixture-of-Experts (MoE) paradigm, MACE utilizes a gating network to classify global descriptors and dynamically activate a single subnetwork per inference, reducing computational cost to the level of small-scale scenes.
To balance sub-network training across experts, we introduce an Auxiliary-Loss-Free Load Balancing (ALF-LB) strategy, which improves both angular and translational accuracy in large-scale localization.
For rendering, MACE's high-precision point cloud output is used to regress 3DGS parameters via a Gaussian prediction head.
Specifically, the point clouds inferred by MACE are serve as Gaussian centers, while features from fully convolutional upsampled maps are integrated to predict the remaining pixel-aligned 3DGS parameters, enabling high-quality rendering from the input image perspective.

To summarize, our main contributions are as follows: (1) We propose MACE, a Mixture-of-Experts-based framework for large-scale scene localization and rendering, which activates only one sub-network per inference to reduce computation without sacrificing accuracy. (2) We introduce an \emph{Auxiliary-Loss-Free Load Balancing} strategy to ensure effective sub-network training without additional loss terms, achieving lower angular and translational errors. (3) We leverage MACE-inferred point clouds with a Gaussian regression head to predict 3DGS parameters, bridging the gap between localization and rendering. (4) Extensive experiments demonstrate both efficient localization and high-quality rendering on large-scale scenes, outperforming state-of-the-art methods.

\section{Related work} \label{sec::related work}

\subsection{Pose Regression and Feature Matching}
Camera pose estimation has traditionally relies on feature detection and matching~\cite{lowe2004distinctive, bo2010kernel}, but such methods suffer from degraded performance in low-overlap scenarios. Learning-based pose regression methods~\cite{kendall2015posenet} directly predict 6-DoF poses from images, offering improved robustness but generally lower accuracy compared to geometry-based approaches. Relative pose regression techniques~\cite{sattler2019understanding,turkoglu2021visual,chen2024map} improve generalization by predicting relative transforms between query and reference images, yet their localization precision remains limited.
Feature matching becomes the dominant paradigm for visual localization~\cite{detone2018superpoint,sarlin2019coarse,sarlin2020superglue}, establishing 2D–3D correspondences between query images and pre-constructed 3D models. To support large-scale environments, many approaches~\cite{arandjelovic2016netvlad,torii201524} employ image retrieval for coarse localization followed by fine-grained matching. However, these methods often incur substantial storage and computational costs due to descriptor-heavy 3D representations. Recent methods such as GoMatch~\cite{zhou2022geometry} and MeshLoc~\cite{panek2022meshloc} reduce memory overhead by matching against scene geometry, but still depend on computationally expensive structure-from-motion pipelines. Even with recent acceleration strategies~\cite{sarlin2020superglue,potje2024xfeat}, feature-based systems still constrain by long mapping times and large memory usage.

\subsection{Large-scale Scene Coordinate Regression}
Scene Coordinate Regression (SCR) methods bypass explicit descriptor matching by directly regressing 3D scene coordinates from 2D image pixels using implicit representations encoded in neural networks~\cite{shotton2013scene,brachmann2017dsac,brachmann2021visual}. Early SCR approaches relied on random forests~\cite{shotton2013scene,valentin2015semanticpaint} or adopted CNN-based architectures~\cite{huang2021vs,wang2024hscnet++}, offering compact map sizes.
ACE~\cite{brachmann2023accelerated} has demonstrated strong performance in small-scale scenarios, achieving rapid training speed by forgoing explicit 3D reconstruction.
Recently, several approaches have been proposed to improve the scalability and performance of SCR in large-scale scenes. These methods often rely on ground truth 3D coordinates and aim to handle large scenes by dividing them into smaller segments, such as spatial regions~\cite{brachmann2019expert}, voxels~\cite{tang2023neumap}, or hierarchical clusters~\cite{li2020hierarchical,wang2024hscnet++}.
However, extending SCR to large-scale scenes remains challenge. The limited capacity of a single network often constrains performance, and existing solutions typically rely on ensembles of specialized subnetworks~\cite{brachmann2019expert, brachmann2023accelerated}, leading to increased computational overhead.
Our work addresses this gap by introducing a MoE architecture that dynamically activates a single subnetwork per inference. This design preserves the efficiency of single-network methods in small scenes while enabling effective scalability to large environments.



\subsection{Feed-Forward Rendering}
Feed-forward methods enable fast inference from sparse views by leveraging large-scale priors and are broadly categorized into NeRF-based and Gaussian-based approaches.
NeRF-based methods~\cite{yu2021pixelnerf, wang2021ibrnet, du2023learning} pioneered the feed-forward rendering paradigm. While Neural Radiance Fields~\cite{mildenhall2021nerf} produce photorealistic results, their rendering speed remains a major limitation.
In contrast, 3D Gaussian Splatting~\cite{kerbl20233d} achieves real-time rendering by replacing expensive volume sampling with rasterized Gaussian primitives. Therefore, Gaussian-based feed-forward extensions, such as GPS-Gaussian~\cite{zheng2024gps}, pixelSplat~\cite{charatan2024pixelsplat}, and MVSplat~\cite{chen2024mvsplat} outperform previous NeRF-based methods. However, they require hours of per-scene optimization in large-scale scenes.
To address this, variants like DepthSplat~\cite{xu2025depthsplat} predict Gaussian parameters from multi-view monocular depth cues, yet still depend on depth supervision.
Alternatively, data-driven approaches such as LGM~\cite{tang2024lgm} utilize large-scale pretraining but incur significant computational overhead.
In this work, we bridges this gap by leveraging accurate geometric priors for feed-forward model with a Gaussian regression head, enabling single-view reconstruction for AR developers.

\begin{figure*}[ht]
	\centering
	\scalebox{1}{\includegraphics[width=1.0\linewidth]{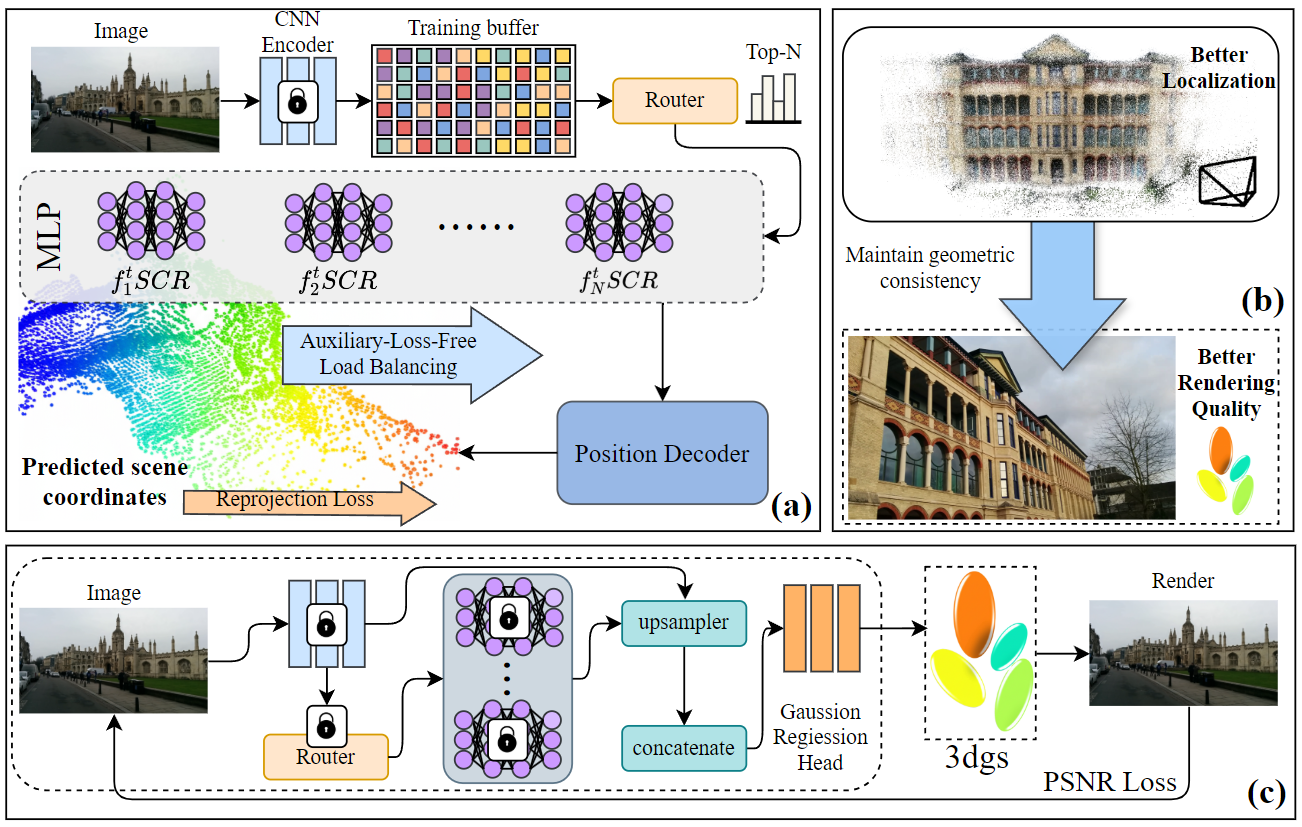}}

	\caption{
	\textbf{Overview of MACE. (a)} Localization Pipeline. Input images are encoded via a pretrained CNN, with local features stored in a training buffer. A router, guided by \textit{Auxiliary-Loss-Free Load Balancing}, selects expert MLPs ($f^t_{\cdot}\text{SCR}$) jointly optimized by reprojection loss to enforce geometric consistency. The MLP output is refined by a \textit{Position Decoder} to generate final coordinate predictions. \textbf{(b)} Localization-rendering cascade. Better localization via the MACE localization pipeline directly improves rendering quality.
	\textbf{(c)}
	The router selects pretrained expert MLPs for coordinate prediction, which are fused with features and point clouds, then processed by a Gaussian head to generate static views optimized with PSNR loss.
	}
	\label{fig:pipeline}
\end{figure*}

\section{Preliminaries}
\subsection{Accelerated Coordinate Encoding}
\label{sec:ace}
SCR methods aim to establish implicit 2D-3D correspondences by predicting a dense scene coordinate map from input images using a convolutional neural network. Given an image patch $p_i$ centered at pixel ($x_i$, $y_i$), the network predicts its corresponding 3D coordinate $z_i$ as:
\begin{equation}
	z_i = f(p_i),
\end{equation}
where $f$ denotes the regression function implemented by the neural network.

Traditionally, SCR models were trained using ground-truth 3D scene coordinates as supervision. However, recent advances have enabled training without ground-truth through the use of a differentiable reprojection loss. Accelerated Coordinate Encoding (ACE)~\cite{brachmann2023accelerated} adopts this unsupervised training paradigm and achieves strong performance on small-scale scenes.
The training is driven by minimizing the reprojection loss over all training views $\{I_i\}_{i=1}^N$:
\begin{equation}
	\arg\min_{w} \sum_{i=1}^{N} \sum_{x_j \in I_i} \ell_{\pi}\left(x_j,  z_j, T_i \right),
\end{equation}
where $w$ represents the learnable parameters of ACE,
and $T_i$ is the camera pose matrix for view $I_i$.
The function $\ell_{\pi}$ is the DSAC$^*$-based reprojection loss ~\cite{brachmann2021visual} that measures the discrepancy between the projected 3D point and its observed 2D location.
By optimizing this objective, ACE effectively learns scene geometry in an end-to-end manner without requiring explicit 3D supervision.

\subsection{Challenges of ACE in Large-Scale Scenes}
While ACE achieves state-of-the-art performance in small-scale scenes, its extension to large-scale environments introduces critical limits stemming from both its core mechanism and practical deployment constraints.

From a theoretical standpoint, ACE relies on an implicit triangulation process by independently regressing each 2D observation to a 3D point through a neural network. Owing to the smooth nature of neural function approximation, similar visual features are encouraged to regress to similar spatial coordinates by minimizing the aggregate reprojection error:
\begin{equation}
	\mathcal{L}_{\text{reproj}} = \sum_{i} \| \pi(\mathbf{X}_i) - \mathbf{X}_i \|^2,
\end{equation}
where $\mathbf{x}_i$ is the 2D observation, $\mathbf{X}_i$ is the regressed 3D coordinate, and $\pi(\cdot)$ denotes the camera projection function.

This approach works well in small scenes with consistent visual descriptors but struggles in large-scale settings.
Specifically, ACE handles large-scale scenes by dividing them into spatial clusters, and a dedicated subnet is trained for each. During inference, all subnets are executed in parallel, and the most confident prediction is selected.
However, these approaches have two key limitations:
\begin{itemize}
	\item \textbf{Suboptimal clustering boundaries:} Scene partitioning by pose proximity may separate visually or geometrically coherent regions, disrupting feature continuity and impairing learning.

	\item \textbf{High inference cost:} Concurrent execution of all subnets incurs substantial computational overhead and scales linearly with cluster count, limiting scalability to larger scenes and real-time applicability.
\end{itemize}

These drawbacks motivate a more robust architecture that ensures feature consistency and computational efficiency without hard scene partitioning.
Additionally, repeated textures and viewpoint variations cause same point exhibit distinct features under varying perspectives. The mismatch between visual similarity and spatial correspondence, leading to significant localization errors.

\section{Methodology}
\subsection{Overview}
In this section, we introduce MACE framework to overcome the limitation of ACE in large-scale settings.
To enhance localization, we introduce an \textit{Auxiliary-Loss-Free Load Balancing} (ALF-LB) strategy for efficient expert selection. Additionally, a \textit{Position Decoding} module is utilized to mitigate unimodal prior bias.
Finally, we integrate the localization component into static-view rendering pipeline, demonstrating that improved localization not only accelerates rendering but also enhances visual quality.

\subsection{MoE for Implicit Global Description}
To address the challenge of modeling global information in large-scale scenes, we propose an implicit global representation framework based on a Mixture-of-Experts architecture. As illustrated in Fig.~\ref{fig:pipeline}, the system dynamically selects a pretrained coordinate regression expert, enabling efficient localization without introducing explicit global descriptors.

\paragraph{Expert Pretraining.}
We first partition the scene into $K$ spatial clusters based on camera pose distribution. For each cluster, we pretrain an expert network $\mathcal{E}_k$ following the ACE ~\cite{brachmann2023accelerated} pipeline:
\begin{equation}
	\mathbf{z}_j = \mathcal{E}_k(\mathbf{f}_j),
\end{equation}
where $\mathbf{f}_j$ is the local feature at pixel $(x_j,y_j)$ and $\mathbf{z}_j$ is the predicted 3D scene coordinate. Each expert learns a local feature-to-coordinate mapping specific to its subregion, ensuring spatial specialization.

\paragraph{Gating Network.}
With experts fixed, we train a gating network $\mathcal{G}$ to predict the most suitable expert for a given image $I_i$. The input to $\mathcal{G}$ is an image-level feature embedding, and the predicted expert index $\hat{k}$ minimizes the reprojection loss of selected coordinates:
\begin{equation}
	\hat{k} = \arg\min_k \sum_{x_j \in I_i} \ell_\pi\left(\mathcal{E}_k(\mathbf{f}_j), \mathbf{T}_i\right),
\end{equation}
where $\mathbf{T}_i$ is the ground-truth pose and $\ell_\pi$ is the DSAC*-based reprojection loss ~\cite{brachmann2021visual}. This training encourages the gating network to learn global spatial distributions from local features.

\paragraph{Joint Optimization.}
After pretraining, we jointly optimize the gating and expert networks. The final coordinate prediction for pixel $(x_j,y_j)$ in image $I_i$ is given by:
\begin{equation}
	\mathbf{z}_j = \mathcal{E}_{\mathcal{G}(I_i)}(\mathbf{f}_j).
\end{equation}
The entire framework is trained end-to-end with reprojection loss.

The advantage of this architecture lies in its implicit encoding of global context. Since local features encode structural and textural cues, their spatial patterns naturally reflect global subregion affiliation.
The gating network learns to leverage this implicit global signal, enabling accurate subregion classification without explicit global descriptors or additional computational overhead.


\subsection{Auxiliary-Loss-Free Load Balancing}
The ALF-LB strategy is introduced to mitigate the expert imbalance issue in traditional MoE architectures, where certain sub-networks are excessively activated while others are rarely utilized.
In contrast to entropy-based auxiliary loss methods, our approach employs a closed-loop bias modulation mechanism that ensures balanced expert activation while maintaining end-to-end differentiability and avoiding conflicting optimization objectives.

\paragraph{Gating with Bias Embedding.} The gating network first computes the raw expert logits via an MLP:
\begin{equation}
	z = \text{MLP}(x),
\end{equation}
where $x$ is the input feature and $z \in \mathbb{R}^K$ are the unnormalized selection logits over $K$ experts. A learnable bias term $b \in \mathbb{R}^K$ is then added to guide expert selection:
\begin{equation}
	\tilde{z} = z + b.
\end{equation}

\paragraph{Differentiable Expert Selection.} We adopt the Gumbel-Softmax trick to enable differentiable sampling from the expert distribution:
\begin{equation}
	\alpha_k = \frac{\exp((\tilde{z}_k + g_k)/\tau)}{\sum_{j=1}^K \exp((\tilde{z}_j + g_j)/\tau)}, \quad g_k \sim \text{Gumbel}(0,1),
\end{equation}
where $\tau$ is the temperature, and $\alpha_k$ denotes the soft assignment weight for expert $k$.

\paragraph{EMA-Based Bias Adjustment.} To ensure long-term load balancing, we maintain a running estimate of expert usage via exponential moving average (EMA):
\begin{equation}
	u_k^{(t)} = \gamma u_k^{(t-1)} + (1 - \gamma)\alpha_k^{(t)},
\end{equation}
where $u_k^{(t)}$ is the usage of expert $k$ at step $t$, and $\gamma$ is the EMA decay rate. The bias $b_k$ is updated as:
\begin{equation}
	b_k \leftarrow b_k - \eta \cdot (u_k^{(t)} - \bar{u}),
\end{equation}
where $\bar{u} = \frac{1}{K} \sum_{j=1}^K u_j^{(t)}$ is the average usage, and $\eta$ is the adjustment rate.

The final output is computed as a weighted sum of expert predictions:
\begin{equation}
	y = \sum_{k=1}^K \alpha_k \cdot f_k(x),
\end{equation}
where $f_k$ is the $k$-th expert sub-network.

This auxiliary-free balancing strategy enhances the spatial specialization of sub-networks, which is crucial in large-scale scene localization. Balanced expert usage ensures that each expert focuses on distinct spatial regions, improving the discriminative quality of global descriptors and mitigating the issue of spatial coverage bias caused by overfitting in dominant experts.

\subsection{Position Decoding}
Previous research~\cite{sattler2019understanding} has shown that the design of the final layer in a convolutional neural network significantly affects the model's prior when regressing spatial positions. Specifically, when the final linear layer outputs a linear combination of weight bases, it restricts the flexibility of position parameterization. This limitation becomes particularly pronounced in scenarios lacking ground-truth scene coordinate supervision, where the model relies on its prior for implicit triangulation.
In the original ACE method, the network predicts an offset relative to a fixed training camera center $c$ in homogeneous coordinates. This design imposes a unimodal prior, causing the predicted positions to cluster around the camera center.

To overcome this limitation, we adopt a more flexible position decoding strategy previously proposed in GLACE~\cite{wang2024glace}. Camera positions in the training set are first grouped into \( k \) clusters via K-Means, producing cluster centers \( \{\mathbf{c}_i\}_{i=1}^{k} \). The final MLP layer then outputs \( k \) logits \( \{s_i\}_{i=1}^{k} \) and an offset vector. A convex combination of the cluster centers is computed utilizing the softmax-normalized logits, replacing the original fixed center \( \mathbf{c} \). This strategy allows for more expressive and adaptive position estimation, as illustrated below:
\begin{align}
	\hat{\mathbf{c}} = \sum_{i=1}^k \frac{e^{s_i}}{\sum_j e^{s_j}} \mathbf{c}_i .
\end{align}
This scheme introduces multimodal characteristics through the dynamic weighting of cluster centers, and the effect of this scheme can be observed in the ablation experiments.

\subsection{Combined with Forward Rendering Pipeline}
Conventional rendering pipelines rely heavily on Structure-from-Motion (SfM) ~\cite{schonberger2016structure} for point cloud priors, but SfM is inefficient in large-scale scenes. Existing 3D reconstruction methods require depth supervision, point clouds, or data-intensive training. In contrast, we propose the first unsupervised feed-forward rendering pipeline based on Gaussian Splatting~\cite{kerbl20233d}, enabling scale-consistent rendering without explicit geometric supervision via a localization-optimized framework.

As illustrated in the Fig.~\ref{fig:pipeline} (c), we first freeze the parameters of the trained gating network and expert sub-networks from our localization framework. Given a large-scale input view, the network infers a per-view feature map $\mathbf{F}$ and sparse point cloud $\mathcal{P} = \{\mathbf{p}_i \in \mathbb{R}^3\}_{i=1}^N$. Then, the feature map $\mathbf{F}$ is up-sampled to a target resolution via fully convolutions. The corresponding point cloud $\mathcal{P}$ is bilinearly interpolated into the same grid, resulting in a dense feature-volume pair $\{\mathbf{F}', \mathbf{P}'\} \in \mathbb{R}^{H \times W \times (C + 3)}$.
Additionally, we fix $\mathbf{P}'$ as spatial anchors and use a fully  convolutional network $\mathcal{F}$ to regress the remaining Gaussian parameters:
\begin{align}
	[\alpha_i, \mathbf{c}_i, \Sigma_i] = \mathcal{F}([\mathbf{F}'_i, \mathbf{P}'_i]),
\end{align}
where $\alpha$, $\mathbf{c}$ and $\Sigma_i$ are the opacity, color, and covariance, respectively. This factorization decouples geometry from appearance, facilitating efficient learning with consistent structure.
The final rendered image $\mathbf{I}_{\text{render}}$ is supervised by a photometric loss against the input image $\mathbf{I}_{\text{gt}}$:
\begin{align}
	\mathcal{L}_{\text{gs}} = (1- \lambda)\mathcal{L}_\text{MSE}(\mathbf{I}_{\text{render}}, \mathbf{I}_{\text{gt}}) + \lambda \mathcal{L}_\text{D-SSIM}.
\end{align}


Our method removes reliance on SfM priors, depth labels, or large datasets. Leveraging localization-enhanced features, the unsupervised feed-forward 3DGS model enables high-fidelity rendering for large-scale rendering.


\section{Experiments}
\subsection{Experimental Setup}
\noindent{\bf{Datasets.}}
We evaluate our MACE on the Cambridge Landmarks dataset~\cite{kendall2015posenet}, which contains extensive outdoor scenes of historic buildings in Cambridge city center. The dataset includes rich sets of mapping and query images, with ground-truth camera poses jointly reconstructed via SfM, providing a reliable benchmark for localization and rendering accuracy.


\noindent{\bf{Baselines.}}
To assess MACE's effectiveness in managing activation map scale for large-scale localization, we compare it with baselines from three paradigms—FM, APR, and SCR—focusing on localization accuracy and computational cost. Key settings and results are summarized in Tab.~\ref{tab:results_cam}.


\noindent{\bf{Metrics.}}
We adopt multi-dimensional metrics to evaluate MACE in large-scale localization and forward rendering tasks. Localization accuracy is measured by the median translation error and rotation error between the predicted and ground-truth poses. Computational efficiency is reflected by the memory footprint of activated map weights. For rendering quality, we assess visual fidelity using Peak Signal-to-Noise Ratio (PSNR), Structural Similarity Index (SSIM)~\cite{1284395}, and Learned Perceptual Image Patch Similarity (LPIPS)~\cite{zhang2018unreasonable}, comparing rendered views against reference frames.

\noindent{\bf{Implementation Details.}}
MACE is implemented in PyTorch, building upon the public ACE codebase~\cite{brachmann2023accelerated}. For standard Cambridge scenes, we train with a batch size of 40K, buffer size of 16M, and 16 epochs on an NVIDIA RTX 3090. For the more complex \emph{GreatCourt} scene, we adopt an enhanced gating network, increase batch size to 160K and buffer size to 64M, extend training to 30 epochs, and utilize an NVIDIA A800. The number of activated sub-networks is dynamically adjusted based on scene complexity. The number of decoder clusters is set to 50, determined via hyperparameter tuning.

When training the Gaussian regression head, we use the AdamW optimizer with a learning rate ranging from 2e-4 to 2e-3, adopting a one cycle learning rate scheduling strategy. To speed up training, the regression head is trained with half-precision floating point weights. All experiments are run on a single NVIDIA A800 GPU. During the 10-minute training, for the training set of a single scene, we conduct 8 epochs with a batch size of 14.
\begin{figure}[t]
	\centering
	\scalebox{1}{\includegraphics[width=1.0\linewidth]{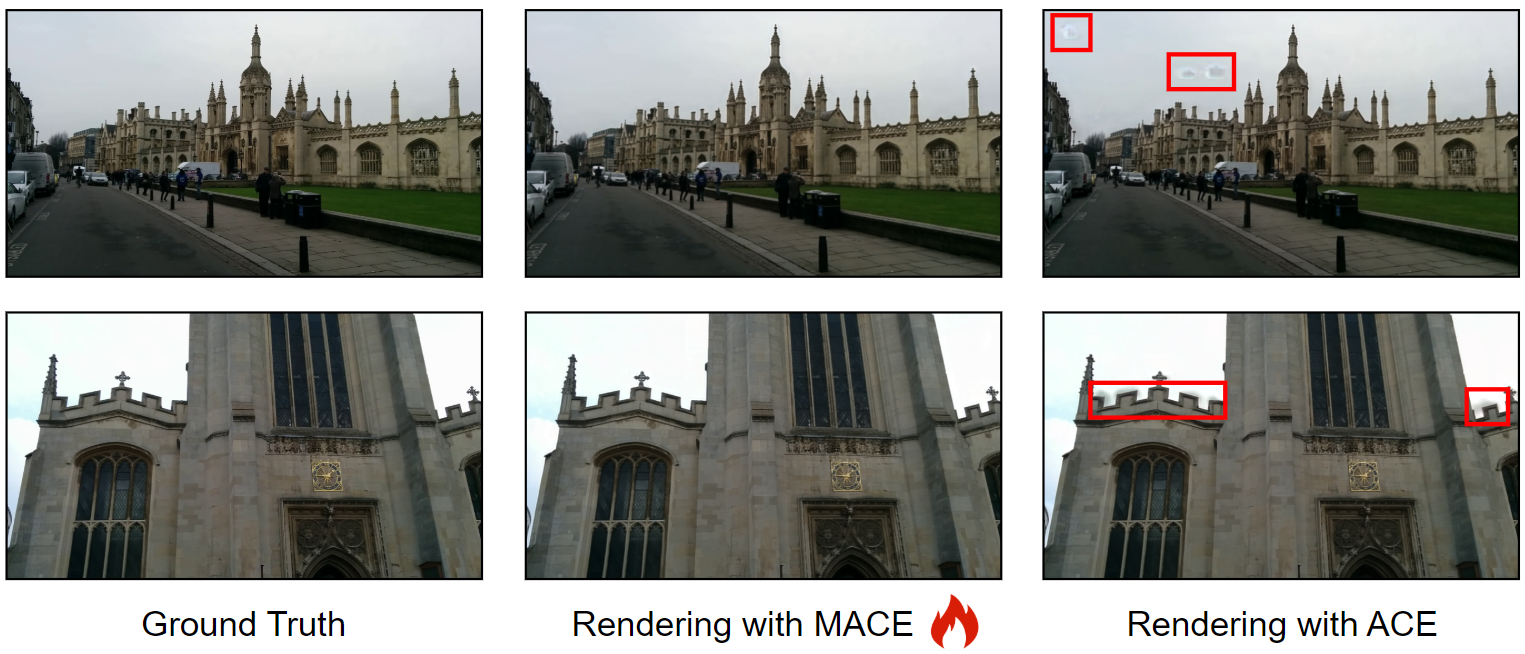}}

	\caption{
		\textbf{Qualitative comparison across different localization frameworks.} Red boxes in ACE highlight artifacts such as misaligned architectural details, emphasizing the superior visual fidelity achieved by MACE through more accurate localization.
	}
	\label{fig:qualitative comparison}
\end{figure}

\begin{figure}[t]
	\centering
	\scalebox{1}{\includegraphics[width=1.0\linewidth]{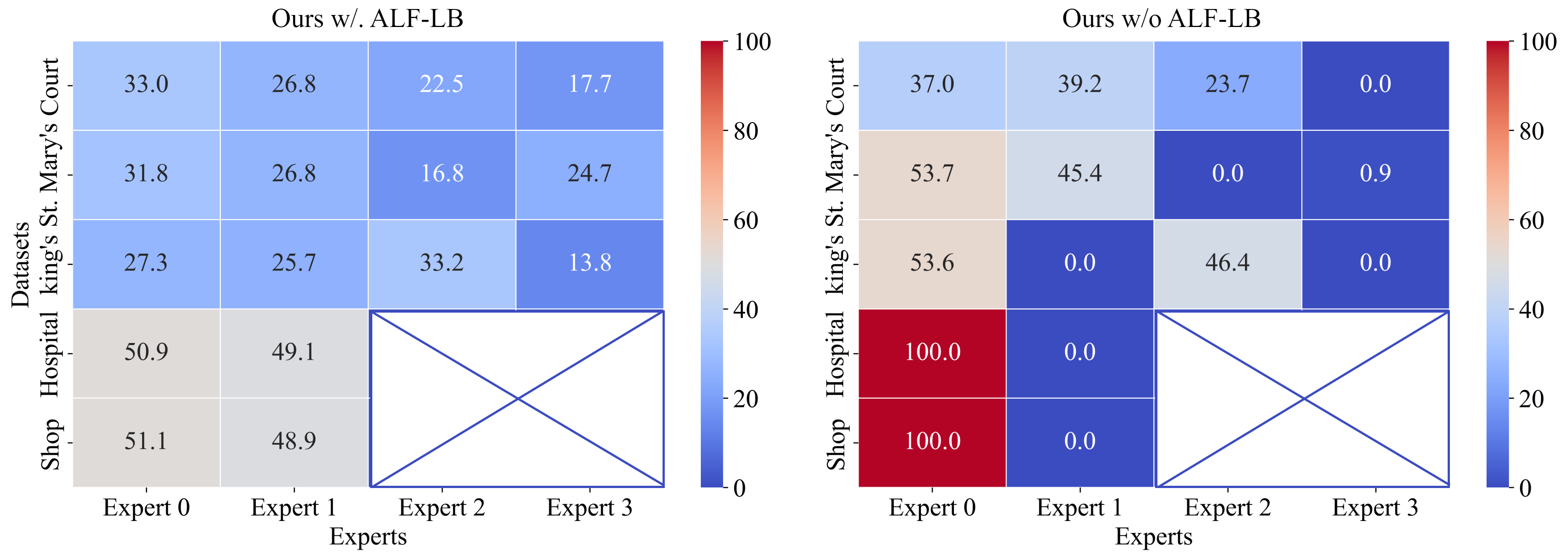}}

	\caption{
		\textbf{Ablation of MoE architecture.} We compare the heatmaps of gating network activations with and without ALF-LB strategy. The left exhibits balanced expert utilization, whereas the right reveals significant imbalance, highlighting ALF-LB's effectiveness in mitigating utilization bias.
	}
	\label{fig:comparision_result}
\end{figure}

\begin{table*}[t]
	\centering
	\footnotesize
	\setlength{\tabcolsep}{4pt}
	\begin{tabular}{clcccccccc}
		\toprule
		 & Method               & \begin{tabular}[c]{@{}c@{}}Mapping w/ Depth\end{tabular} & \begin{tabular}[c]{@{}c@{}}Map Size\end{tabular}
		 & GreatCourt           & Kings                                                    & Hospital                                         & Shop            & StMary          & \begin{tabular}[c]{@{}c@{}}Average (cm / $^\circ$)\end{tabular}                                                     \\
		\midrule
		\multirow{5}{*}{\rotatebox{90}{FM}}
		 & AS (SIFT)            & No                                                       & $\sim$200MB                                      & 24/0.1          & 13/0.2          & 20/0.4                                                          & 4/0.2          & 8/0.3          & 14/0.2          \\
		 & hLoc (SP+SG)         & No                                                       & $\sim$800MB                                      & 16/0.1          & 12/0.2          & 15/0.3                                                          & 4/0.2          & 7/0.2          & 11/0.2          \\
		 & pixLoc               & No                                                       & $\sim$600MB                                      & 30/0.1          & 14/0.2          & 16/0.3                                                          & 5/0.2          & 10/0.3         & 15/0.2          \\
		 & GoMatch              & No                                                       & $\sim$12MB                                       & N/A             & 25/0.6          & 283/8.1                                                         & 48/4.8         & 335/9.9        & N/A             \\
		 & HybridSC             & No                                                       & $\sim$1MB                                        & N/A             & 81/0.6          & 75/1.0                                                          & 19/0.5         & 50/0.5         & N/A             \\
		\midrule
		\multirow{2}{*}{\rotatebox{90}{APR}}
		 & PoseNet17            & No                                                       & 50MB                                             & 683/3.5         & 88/1.0          & 320/3.3                                                         & 88/3.8         & 157/3.3        & 267/3.0         \\
		 & MS-Transformer       & No                                                       & $\sim$18MB                                       & N/A             & 83/1.5          & 181/2.4                                                         & 86/3.1         & 162/4.0        & N/A             \\
		\midrule
		\multirow{3}{*}{\rotatebox{90}{\shortstack{SCR                                                                                                                                                                                                                                                  \\ w/ \\ Depth}}}
		 & DSAC* (Full)         & Yes                                                      & 28MB                                             & 49/0.3          & 15/0.3          & 21/0.4                                                          & 5/0.3          & 13/0.4         & 21/0.3          \\
		 & SANet                & Yes                                                      & $\sim$260MB                                      & 328/2.0         & 32/0.5          & 32/0.5                                                          & 10/0.5         & 16/0.6         & 84/0.8          \\
		 & SRC                  & Yes                                                      & 40MB                                             & 81/0.5          & 39/0.7          & 38/0.5                                                          & 19/1.0         & 31/1.0         & 42/0.7          \\
		\midrule

		\multirow{4}{*}{\rotatebox{90}{SCR}}
		 & DSAC* (Full)         & No                                                       & 28MB                                             & 34/0.2          & 18/0.3          & 21/0.4                                                          & 5/0.3          & 15/0.6         & 19/0.4          \\
		 & DSAC* (Tiny)         & No                                                       & 4MB                                              & 98/0.5          & 27/0.4          & 33/0.6                                                          & 11/0.5         & 56/1.8         & 45/0.8          \\
		 & ACE                  & No                                                       & 4MB                                              & 43/0.2          & 28/0.4          & 31/0.6                                                          & 5/0.3          & 18/0.6         & 25/0.4          \\
		 & Poker (ACE$\times$4) & No                                                       & 16MB                                             & 28/0.1          & 18/0.3          & 25/0.5                                                          & 5/0.3          & \textbf{9/0.3} & 17/0.3          \\

		\midrule
		\midrule
		\multirow{3}{*}{\rotatebox{90}{MACE}}
		 & Ours w/o ALF-LB      & No                                                       & 4.25$\sim$5.26MB                                             & 32/0.2          & 20/0.3          & 28/0.5                                                          & 6/0.3          & 14/0.4         & 20/0.3          \\

		 & Ours w/o Decoder     & No                                                       & 4.25$\sim$5.26MB                                             & 27/0.2          & 18/0.3          & 21/0.5                                                          & 5/0.3          & 11/0.4         & 16/0.3          \\

		 & Full model           & No                                                       & 4.25$\sim$5.26MB                                 & \textbf{24/0.2} & \textbf{15/0.3} & \textbf{19/0.4}                                                 & \textbf{5/0.2} & \textbf{9/0.3} & \textbf{14/0.3} \\
		\bottomrule
	\end{tabular}
	\caption{\textbf{Cambridge Landmarks~\cite{kendall2015posenet} Results.}
		Median translation and rotation errors (cm / $^\circ$).
		\textbf{Bold} indicates best performance in SCR.
	}
	\label{tab:results_cam}
\end{table*}

\begin{table}[h]
\centering
\caption{Mapping Time Across Scenes}
\begin{tabular}{lcc}
\toprule
Scene & GPU Configuration & Mapping Time \\
\midrule
GreatCourt & A800x1 & 30min \\
KingsCollege & RTX 3090x1 & 30min \\
OldHospital & RTX 3090x1 & 20min \\
ShopFacade & RTX 3090x1 & 20min \\
StMarysChurch & RTX 3090x1 & 30min \\
\bottomrule
\end{tabular}
\label{tab:mapping_time}
\end{table}

\begin{figure}[t]
	\centering
	\scalebox{1}{\includegraphics[width=1.0\linewidth]{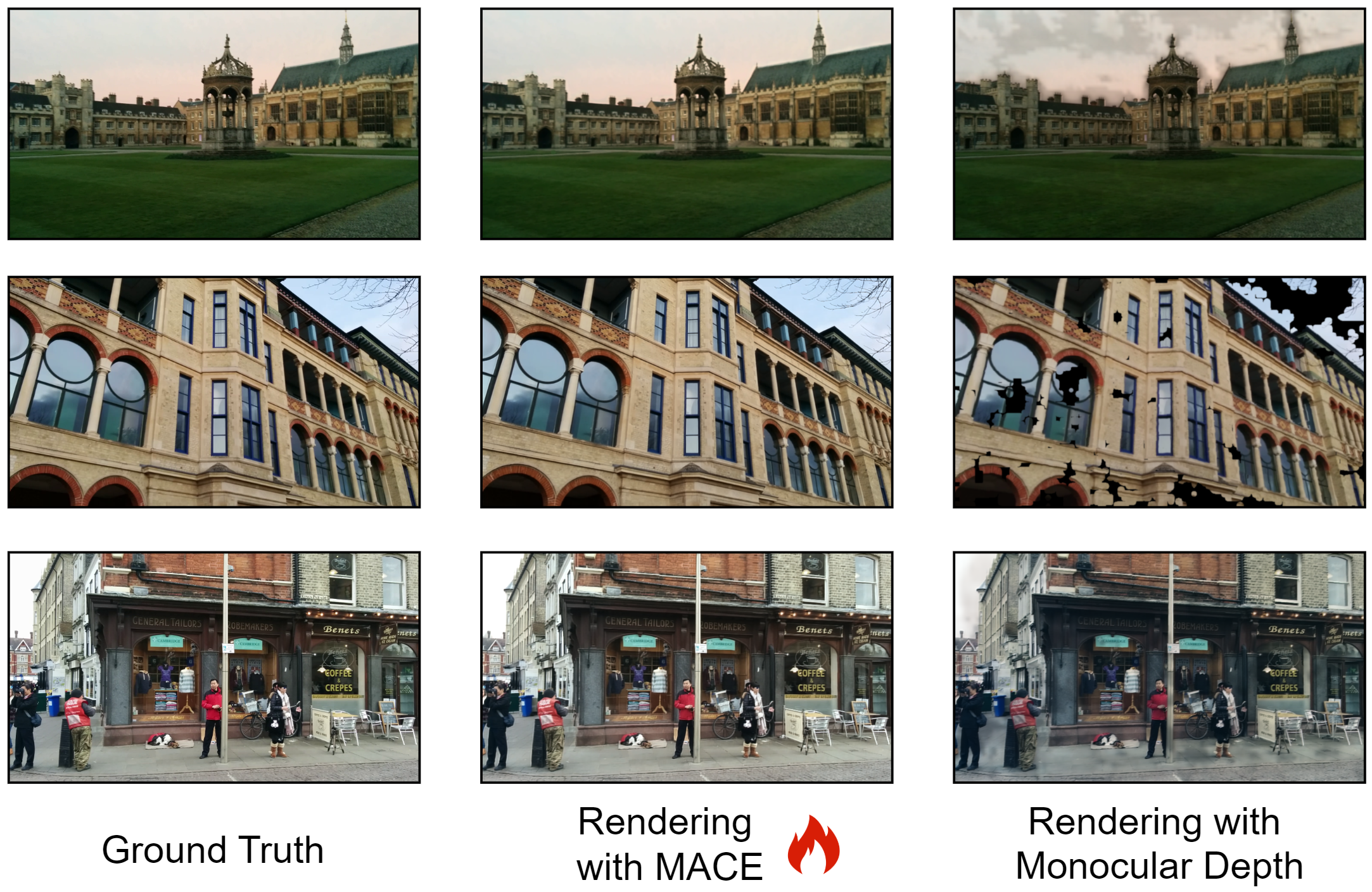}}

	\caption{
		\textbf{Ablation Comparison of Geometric Priors in AR Rendering.} Renderings with monocular depth exhibit severe artifacts like fragmented structures and blackened regions, while MACE - generated renderings closely match the ground truth, demonstrating the superiority of our SCR - based geometric prior in ensuring high - fidelity 3DGS rendering for AR static view tasks.
	}
	\label{fig:ablation_render}
\end{figure}

\subsection{Localization and Rendering Results}
\noindent{\bf{Localization.}}
As shown in Tab.~\ref{tab:results_cam}, our method significantly outperforms the state-of-the-art SCR methods and approaches the accuracy of FM methods. More importantly, compared with the current leading ACE method, our approach achieves better accuracy while requiring only an activation weight comparable to that of a single sub-network in ACE. This indicates that our method not only improves precision but also enhances computational efficiency by leveraging activation weights more effectively.

Beyond localization accuracy and parameter efficiency, we further evaluate the practical deployment feasibility of our method by analyzing its mapping efficiency. Experiments use A800x1 and RTX 3090x1 GPUs.
As shown in Tab.~\ref{tab:mapping_time}, similar to ACE, MACE can be trained on a single GPU, ensuring accessibility and practicality in deployment. While maintaining a comparable training time to ACE, MACE achieves significantly higher precision—striking a favorable balance between efficiency and accuracy that underscores its superiority in scene localization.


\noindent{\bf{Rendering Results.}}
As shown in Tab.~\ref{tab:rendering}, MACE achieves an average PSNR of 34.15\,dB, a truly excellent performance that stands out in the field. Visual comparisons in Fig.~\ref{fig:qualitative comparison} further confirm that MACE reduces distortions present in ACE-based reconstructions. This demonstrates that improved localization leads to better rendering quality.

\begin{figure}[t]
	\centering
	\scalebox{1}{\includegraphics[width=1.0\linewidth]{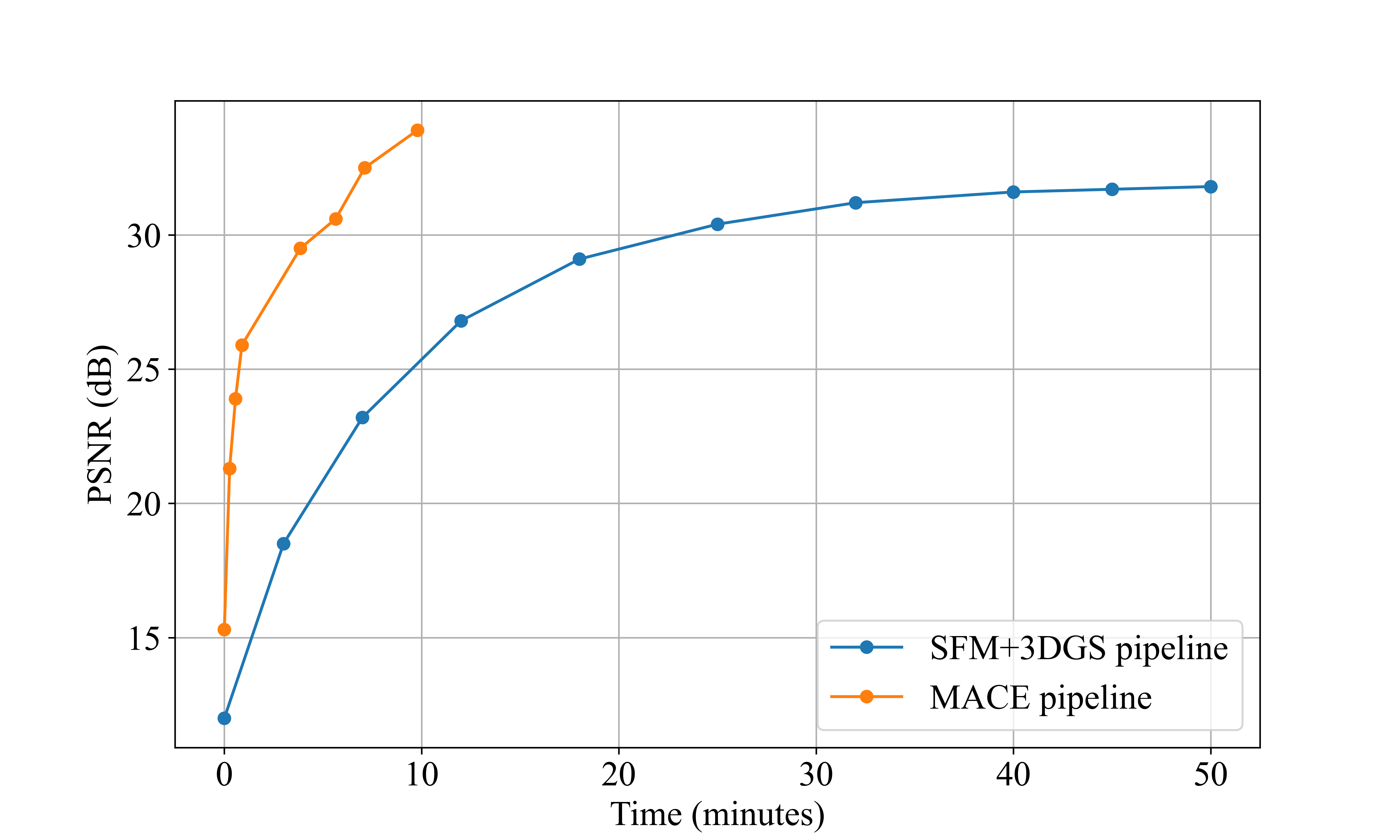}}

	\caption{
		\textbf{Training Time vs. PSNR.} Comparison with SFM+3DGS on the Cambridge dataset shows MACE achieves higher PSNR in 10 minutes than SFM+3DGS does in 50 minutes, highlighting its efficiency from geometric constraints.
	}
	\label{fig:PSNR_time}
\end{figure}

To highlight our approach’s advantage, we compare training time vs. PSNR with the traditional SFM+3DGS pipeline (Fig.~\ref{fig:PSNR_time}). Our pipeline converges to higher rendering quality in under 10 minutes than SFM+3DGS achieves in 50 minutes, underscoring its efficiency and effectiveness in leveraging geometric constraints for better rendering.


\subsection{Ablation Study}
To evaluate the efficacy of MACE in large-scale localization tasks, we conduct ablation experiments on the Cambridge Landmarks dataset to assess two key components of MACE. As shown in Tab.~\ref{tab:results_cam}, removing the ALF-LB strategy leads to uneven expert utilization (Fig.~\ref{fig:comparision_result}) and significantly degrades localization accuracy. Additionally, replacing our decoder with ACE's single-mode prior results in an average increase of 3\,cm in translation error. These findings confirm the importance of both components in achieving accurate and stable scene localization.

For downstream large-scale AR static view rendering, we extend ablation analysis to compare our SCR - based geometric prior with monocular depth - derived priors. We pioneer an unsupervised, data - agnostic SCR - based fast - training method, providing 3DGS with a scale - consistent geometric prior. In experiments, we contrast it with using ZoeDepth (monocular depth model) ~\cite{bhat2023zoedepth} to generate priors via depth map prediction and unprojection. Comparative visuals in Fig.~\ref{fig:ablation_render} show monocular depth - prior renderings have severe defects like fragmented structures and misalignments, unlike MACE. Our SCR prior encodes geometric consistency, avoiding monocular depth estimation errors (e.g., occlusions, texture - less areas), validating its superiority.

\begin{table}
	\caption{ \textbf{Quantitative rendering results on Cambridge Landmarks.}}
	\centering
	\footnotesize
	\setlength{\tabcolsep}{1pt}
	\newcolumntype{Y}{>{\centering\arraybackslash}X}
	\begin{tabularx}{0.999\linewidth}{l||YYYY}
		\toprule
		Scene
		                              & \multicolumn{1}{c}{ PSNR(dB) $\uparrow$} & \multicolumn{1}{c}{SSIM $\uparrow$} & \multicolumn{1}{c}{LPIPS $\downarrow$} & \multicolumn{1}{c}{Time $\downarrow$} \\
		\midrule
		\scriptsize \emph{Hospital}   & 32.56                                    & 0.9729                              & 0.0689                                 & 590s                                  \\
		\scriptsize \emph{King}       & 32.70                                    & 0.9641                              & 0.0940                                 & 609s                                  \\
		\scriptsize \emph{GreadCourt} & 34.12                                    & 0.9722                              & 0.1101                                 & 614s                                  \\
		\scriptsize \emph{Shop}       & 35.14                                    & 0.9809                              & 0.0509                                 & 587s                                  \\
		\scriptsize \emph{StMary}     & 36.24                                    & 0.9799                              & 0.0722                                 & 607s                                  \\
		\scriptsize Average           & 34.15                                    & {0.9740}                            & {0.0792}                               & 601s                                  \\

		\midrule
		\bottomrule
	\end{tabularx}
	\label{tab:rendering}
\end{table}

\section{Conclusion}
We propose MACE, a novel framework for efficient large-scale scene localization and rendering. By introducing auxiliary-loss-free load balancing and an enhanced position decoding module, MACE achieves both accurate localization and efficient computation. Extensive evaluations on Cambridge Landmarks dataset demonstrate that MACE significantly reduces pose errors while maintaining compact activation maps. Furthermore, the integration with 3D Gaussian Splatting enables high-fidelity rendering, highlighting its potential for real-time AR applications on resource-constrained devices. MACE establishes a scalable and accurate paradigm for Large-scale
scene localization and rendering.







\bibliographystyle{IEEEtran}
\bibliography{ref}


\end{document}